\documentclass[preprint,12pt]{elsarticle}

\usepackage{amssymb}
\usepackage{amsthm}
\usepackage[fleqn]{amsmath}
\usepackage{mathtools}
\usepackage{mathrsfs}
\usepackage{graphicx}
\usepackage{subcaption}
\usepackage[space]{grffile}
\usepackage{url}
\usepackage{booktabs}
\usepackage{multirow}
\usepackage{array}
\usepackage{enumitem}
\usepackage[ruled, vlined, linesnumbered]{algorithm2e}
\usepackage{placeins}
\usepackage{float}

\usepackage[pagewise, mathlines]{lineno}

\usepackage[colorlinks]{hyperref}
\hypersetup{
  citecolor = {blue}
}

\usepackage{geometry}
\begin{document}

\begin{frontmatter}

\title{Ramp Metering Control via Hybrid State Deep Reinforcement Learning in Partially Observable Connected Vehicle Environments}

\author[uge,estin]{Youcef Mehamlia\corref{cor1}}
\ead{youcef.mehamlia@univ-eiffel.fr}

\author[uge]{Nadir Farhi}

\author[estin]{Meriem Bouali}

\cortext[cor1]{Corresponding author.}

\address[uge]{Cosys-Grettia, Univ Gustave Eiffel, F-77454 Marne-la-Vallee, France.}

\address[estin]{Laboratoire LITAN, ESTIN, Amizour, Algeria.}


\begin{abstract}
Freeway on-ramp merges are major sources of congestion, causing significant economic and environmental costs. While Deep Reinforcement Learning (DRL) offers a promising solution for ramp metering, existing approaches rely primarily on aggregated macroscopic data. Connected vehicles (CVs) provide vehicle-level observations that can complement aggregate traffic measurements, but their limited penetration produces incomplete microscopic information. This paper proposes a hybrid observation representation combining macroscopic traffic measurements with a two-channel grid encoding observed CV presence and speed. A Dueling Double Deep Q-Network processes these inputs to select ramp-metering green durations. The controller is trained under varying traffic demands and CV penetration rates and evaluated against ALINEA and macroscopic-only DRL variants in SUMO. Across 50 matched evaluation scenarios, the hybrid controller under partial CV visibility reduces the reported total travel time by 11.4\% and mean spillback duration by 84.9\% relative to ALINEA. Evaluating the same trained policy with full CV visibility yields a further travel-time reduction of approximately 1.6\%. Analysis across penetration rates suggests that the performance gap decreases as microscopic observations become more complete. These results support the use of complementary macroscopic and sparse microscopic observations for learning-based ramp metering. The source code implementation of the model is available at: \url{https://github.com/youcefMehamlia/Multimodal-DRL-RMC}
\end{abstract}

\end{frontmatter}

\section{INTRODUCTION}
\label{sec:intro}

Traffic congestion is a major challenge in modern cities. It wastes time \cite{inrix2024}, increases fuel use, and raises emissions \cite{epa2023}. These emissions harm air quality, accelerate climate change, and threaten public health \cite{ipcc2022, who2021}. Much of this congestion starts at freeway on-ramps. When too many vehicles merge at once, they disrupt the mainline traffic flow. This triggers a sudden drop in speed and highway capacity. This phenomenon is known as the \emph{capacity drop} \cite{kerner2004, wang2023_capacitydrop}. Once flow breaks down, recovering free-flow conditions is difficult. Ramp metering solves this by using traffic signals to control the merging rate. Effective metering has two competing goals. It must protect mainline efficiency by preventing the capacity drop. It must also manage ramp queues to prevent spillback onto city streets \cite{treiber2013}.

Early ramp metering relied on fixed-time control. This method uses pre-set rates based on historical data \cite{wattleworth1965}. However, it cannot adapt to real-time traffic changes. This limitation led to local feedback controllers like ALINEA, which adjust the rate to maintain a target mainline occupancy \cite{papageorgiou1991}. Field evaluations have demonstrated the effectiveness of these isolated feedback strategies \cite{HAJSALEM2012378}. To manage larger networks, researchers developed coordinated strategies, such as HERO, to handle multiple ramps and upstream storage \cite{papamichail2010_heuristic}. Advanced coordinated systems also use multi-objective optimization to balance traffic efficiency and safety \cite{HFL16}. Furthermore, field tests show that combining coordinated metering with other physical measures, such as hard shoulder running, significantly improves highway performance \cite{HAJSALEM20141002}. Finally, predictive models, like model predictive control (MPC), emerged to compute optimal rates using traffic forecasts \cite{hegyi2005model, bellemans2006model}.

Connected vehicles (CVs) provide an additional source of traffic information through transmitted positions and speeds \cite{shladover2018connected}. These observations can complement aggregate detector measurements by preserving localized spatial information around the merge. However, only a fraction of vehicles may transmit such data, resulting in incomplete microscopic observations. In this study, CV penetration refers to the availability of vehicle-level measurements, rather than to the proportion of automated vehicles. The resulting control problem therefore concerns ramp metering under partial microscopic observation.

To handle dynamic traffic, control strategies have shifted toward learning-based methods. Reinforcement learning (RL) allows controllers to learn control policies through interaction with the environment \cite{Sutton2018}. Early methods used simple algorithms like tabular Q-learning \cite{Ivanjko2015, fares2014freeway}. Deep reinforcement learning (DRL) advanced this field using neural networks. Algorithms like Deep Q-Network (DQN) and Deep Deterministic Policy Gradient (DDPG) successfully handle continuous, complex traffic states \cite{belletti2018expert_itsc}. Furthermore, multi-agent reinforcement learning (MARL), using algorithms like QMIX, enables the simultaneous control of multiple ramps across a network \cite{Tan2020, rashid2018qmixmonotonicvaluefunction}.

Aggregate traffic measurements provide a compact description of network conditions but do not retain individual vehicle locations. Richer spatial observations have already been investigated for DRL ramp metering: Liu et al.~\cite{liu2021video} use vehicle-position matrices extracted from traffic video. The present work instead examines the combination of aggregate measurements with sparse vehicle-level observations available from CVs. Limited CV penetration leaves some vehicles unobserved, creating a partially observable control problem \cite{Kaelbling1998}. The research question is whether these incomplete microscopic observations complement macroscopic measurements and how the performance of a fixed learned controller changes with CV penetration.

To address this problem, we propose a hybrid-state DRL
ramp-metering controller that combines aggregate traffic
measurements with sparse CV observations. Macroscopic
measurements describe the overall traffic conditions, while
a spatial grid preserves the presence and speeds of observed vehicles near the merge. A two-stream neural network processes these complementary inputs within a Dueling Double DQN controller. The evaluation examines both the benefit of this representation relative to macroscopic-only inputs and the effect of CV penetration on the performance of the same trained hybrid policy.

The main contributions are:
\begin{enumerate}
    \item A hybrid observation representation combining macroscopic traffic measurements with a two-channel spatial grid of observed CV presence and speed.

    \item A ramp-metering controller based on Dueling Double DQN that processes the heterogeneous inputs through separate feature streams and selects discrete
    green-phase durations.

    \item A simulation-based evaluation using matched traffic scenarios, macroscopic-state ablations, and a comparison of the same hybrid policy under partial and full CV visibility to examine the effect of microscopic observation availability.
\end{enumerate}

The rest of the paper is organized as follows. Section~\ref{sec:method} presents the proposed methodology; section~\ref{sec:exp} details the experimental setup; section~\ref{sec:res} discusses the results; and finally, section~\ref{sec:conc} concludes the paper and outlines future work.

\section{METHODOLOGY}
\label{sec:method}

This section presents the proposed Deep Reinforcement Learning (DRL) framework for ramp metering under mixed traffic conditions. The main contribution of this work is a hybrid state representation that integrates macroscopic network-level measurements with microscopic, grid-based telemetry from connected vehicles.

\subsection{Control Problem Formulation}

We formulate ramp metering with incomplete CV observations
as a partially observable control problem. The controller
does not observe the complete traffic state. Instead, at
decision step $t$, it receives the constructed observation
\begin{equation}
o_t =
\left(
\mathbf{s}_{\text{macro}}(t),
\mathbf{S}_{\text{micro}}(t)
\right),
\end{equation}
where $\mathbf{s}_{\text{macro}}(t)$ contains aggregate
traffic measurements, ramp queue information, and the
previous action, while $\mathbf{S}_{\text{micro}}(t)$
encodes the observed CVs.

We use a policy based on the current constructed observation, without an explicit belief-state estimator or recurrent memory. Including aggregated measurements and the previous action provides limited temporal context but does not establish the Markov property. The controller seeks to maximize the expected discounted return:
\begin{equation}
J(\pi)=
\mathbb{E}_{\pi}
\left[
\sum_{t=0}^{\infty}\gamma^t r_t
\right].
\end{equation}
During evaluation, actions are selected greedily:
\begin{equation}
a_t = \arg\max_{a\in\mathcal{A}} Q_{\theta}(o_t,a).
\end{equation}
The selected action determines the ramp-meter green duration
within the control cycle $T_c$.

The following subsections detail the construction of the hybrid observation representation, the neural network architecture, and the reward function guiding the learning process.

\subsection{Hybrid State Representation}

To capture both macroscopic and microscopic traffic dynamics, we construct the hybrid observation $o_t$ from a macroscopic vector $\mathbf{s}_{\text{macro}}(t)$ and a microscopic grid tensor $\mathbf{S}_{\text{micro}}(t)$.This composite representation provides the agent with a comprehensive view of the traffic environment.

The macroscopic vector provides a condensed summary of network operations, functioning similarly to traditional loop detector data. It is a 14-element vector where all variables are normalized to the range $[0, 1]$ to ensure stable neural network training. The features are grouped into aggregate mainline conditions, critical lane specifics, ramp status, and action history, defined formally as:
\begin{equation}
\begin{aligned}
\mathbf{s}_{\text{macro}}(t)
=\operatorname{concat}\Big(
&[\hat{q}_{i,t},\hat{o}_{i,t},\hat{v}_{i,t}]_{i\in\mathcal{I}},
[\hat{l}_{q,t},\hat{a}_{t-1}]
\Big).
\end{aligned}
\end{equation}
where
$\mathcal{I}=
\{\text{upstream},\text{merge},
\text{lane0-upstream},\text{lane0-merge}\}$.

Here, $\hat{q}$, $\hat{o}$, and $\hat{v}$ denote normalized flow, occupancy, and speed for the upstream and merge regions and their right-most lane (Lane 0), where localized merge interactions may trigger breakdown. The vector also includes normalized ramp queue length, $\hat{l}_{q,t}=l_q(t)/l_{q,\max}$, and previous action, $\hat{a}_{t-1}=T_g(a_{t-1})/T_c$.

\begin{figure}[htbp]
\centering
\includegraphics[width=\linewidth]{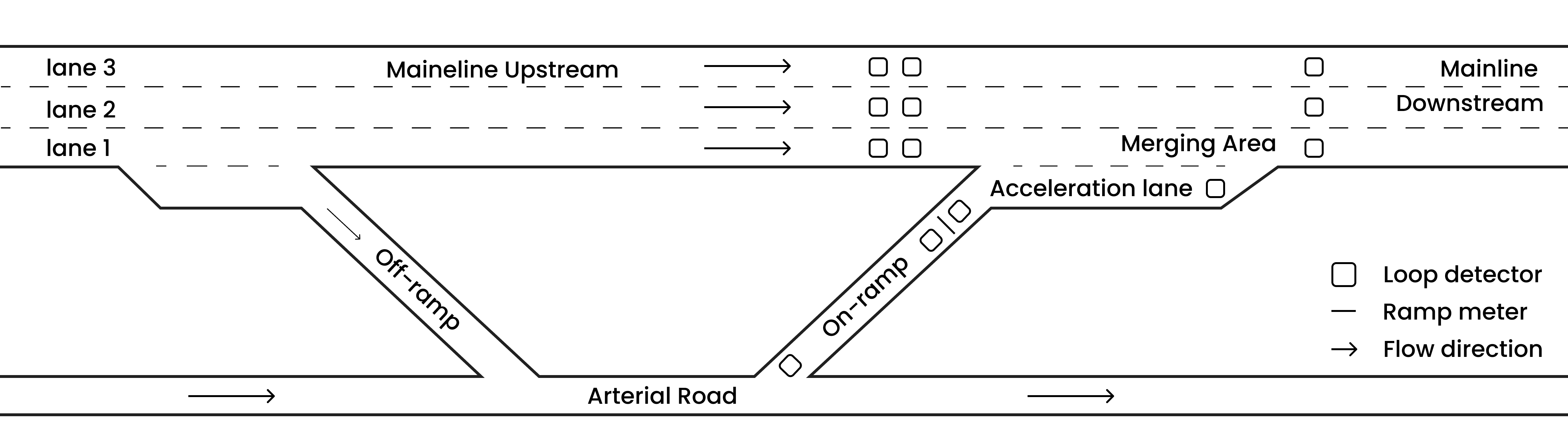}
\caption{Freeway bottleneck and on-ramp configuration.}
\label{fig:network_schematic}
\end{figure}

\begin{figure}[htbp]
\centering
\includegraphics[width=\linewidth]{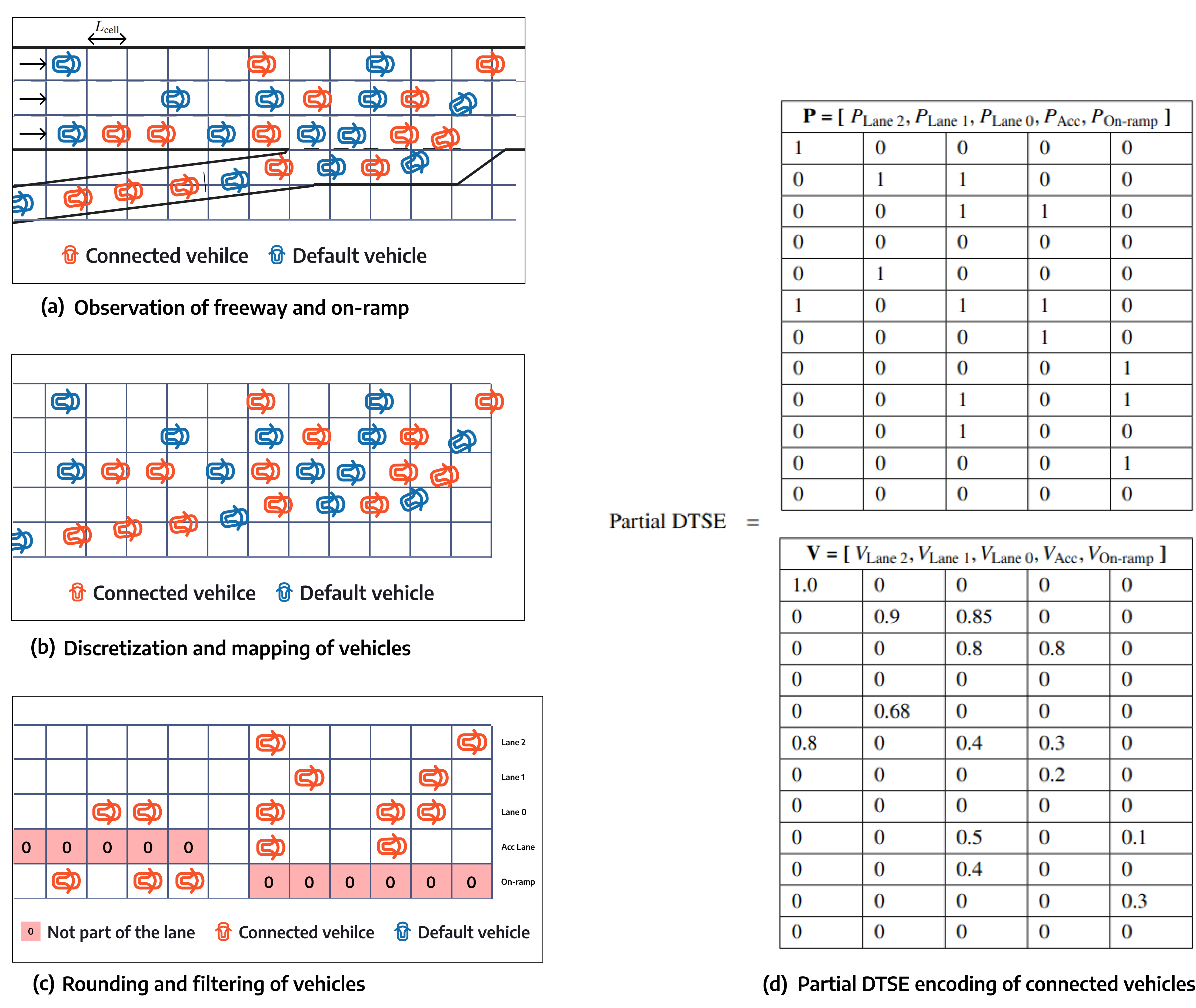}
\caption{Construction of the microscopic CV grid: (a) observation region, (b) spatial discretization, (c) removal of non-connected vehicles, and (d) two-channel encoding of CV presence and normalized speed.}
\label{fig:grid_partial_view_final}
\end{figure}

The microscopic component represents CVs in the merge area as a spatial grid. We discretize a freeway segment of length $l_s$ into an $H \times W$ grid. The number of rows $H$ corresponds to the number of cells along the freeway segment, computed as $H = l_s / L_{\text{cell}}$, where each cell length is $L_{\text{cell}} = \text{Length}_{\text{veh}} + \text{Gap}_{\text{min}} + \text{Buffer}_{\text{safety}}$. The number of columns $W$ equals the total number of lanes in the observation range window $l_r$, including the on-ramp and all freeway lanes within this range. We index the columns inversely, such as column 0 corresponds to the freeway lane furthest from the ramp, and subsequent columns move toward the ramp and the acceleration lane. This arrangement captures the spatial structure of the merge area and aligns with the feature channels (see Fig.~\ref{fig:grid_partial_view_final}).

Only CVs within the observation region are retained, yielding $\mathbf{S}_{\text{micro}}\in\mathbb{R}^{2\times27\times5}$. The first channel is a binary presence mask and the second stores normalized CV speeds. The presence channel distinguishes stopped vehicles from empty cells, since both otherwise have zero speed. Non-connected vehicles remain invisible, producing a sparse representation complemented by the macroscopic measurements.

For storage and transfer to the network, the two components of the observation are serialized as a single vector:
\begin{equation}
o_t = \text{concat}\Big( \mathbf{s}_{\text{macro}}(t), \, \text{flatten}\big(\mathbf{S}_{\text{micro}}(t)\big) \Big)
\end{equation}
The network separates these components and restores the microscopic tensor shape before convolutional processing.

\subsection{Controller and DRL Algorithm}

At each control interval $T_c$, the agent selects an action $a_t$ corresponding to a discrete ramp-meter green duration $T_g$, as listed in Table~\ref{tab:action_map}.

\begin{figure}[htbp]
\centering
\includegraphics[width=\linewidth]{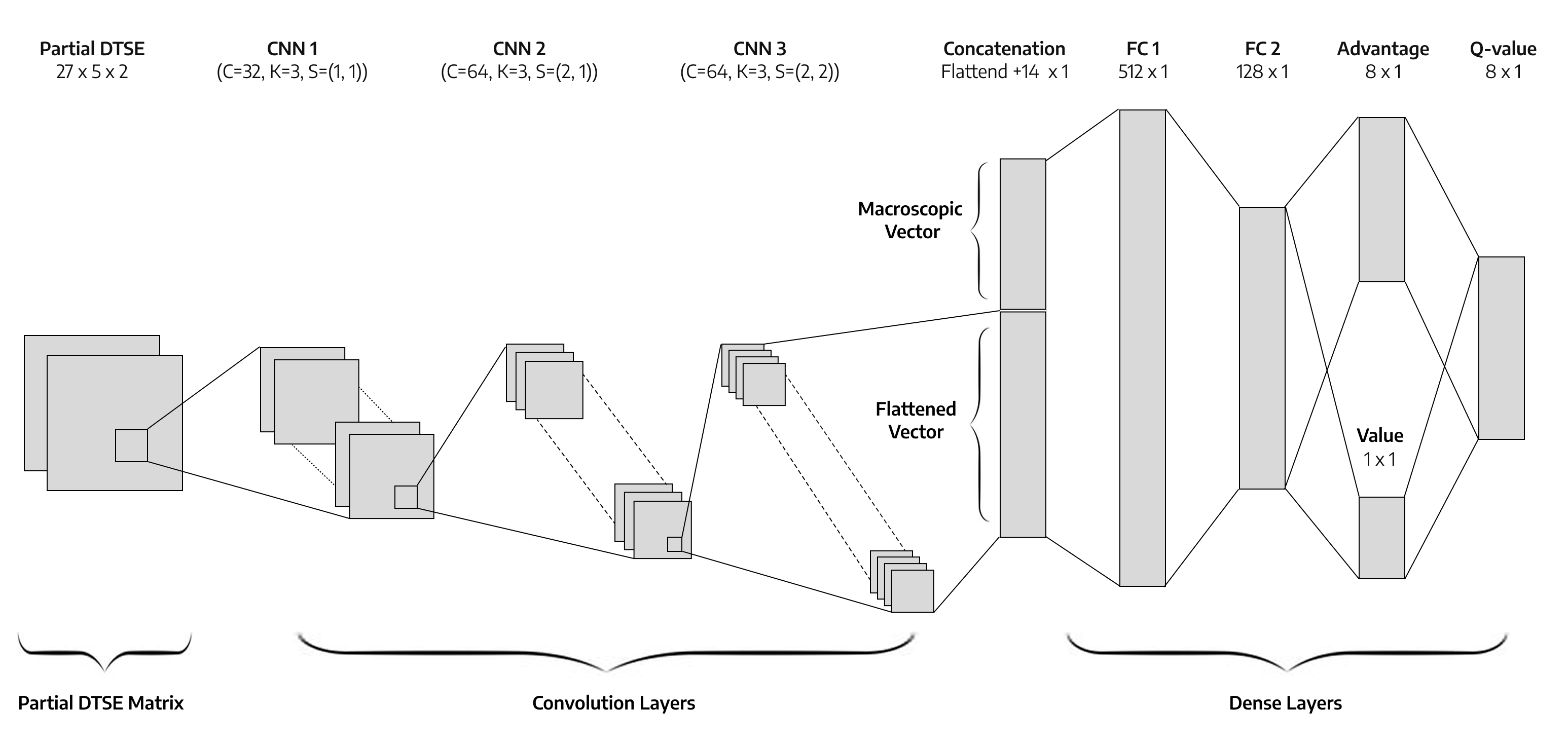}
\caption{Hybrid two-stream architecture combining CNN-based microscopic features and fully connected macroscopic features before the Dueling heads.}
\label{fig:ddqn_architecture}
\end{figure}

To process the hybrid observation, we designed an asymmetric multimodal Two-Stream neural network. The microscopic stream utilizes a multilayer Convolutional Neural Network (CNN) to extract spatial features from the micro state grid $\mathbf{S}_{\text{micro}}(t)$. Concurrently, the macroscopic stream uses a fully connected layer to process the macro state vector $\mathbf{s}_{\text{macro}}(t)$, preserving raw aggregate context. The flattened CNN output is then concatenated with the macroscopic features through a late fusion process, passing through additional shared fully connected layers (Fig.~\ref{fig:ddqn_architecture}).

The network is trained using the Dueling Double Deep Q-Network (Dueling Double DQN) algorithm, chosen to handle high-dimensional state spaces while addressing two major flaws in standard Q-learning. First, Double DQN mitigates overestimation bias by decoupling action selection from action evaluation using a target network, resulting in more accurate value estimates. Second, the Dueling architecture explicitly separates the network heads into two streams, one estimating the observation value $V(o)$ and another estimating the advantage $A(o,a)$ of each action. This decomposition allows the network to represent observation value separately from differences between candidate actions.

\subsection{Reward Function}

The reward balances mainline traffic efficiency against ramp queue accumulation. At each decision step, it combines normalized speeds with penalties for occupancy, ramp queue length, and proximity to ramp storage capacity:
\begin{equation}
\begin{aligned}
r_t ={}&
1.5\hat{v}_{m,t}
+\hat{v}_{u,t}
+0.5\hat{v}_{d,t}
-2\hat{o}_{m,t}
-\hat{o}_{u,t}
-\hat{l}_{q,t}
-20C_{\text{spillback}}.
\end{aligned}
\end{equation}
The subscripts $m$, $u$, and $d$ denote the merge, upstream, and downstream zones, respectively. Speeds are normalized by the reference free-flow speed, occupancy by its normalization constant, and queue length by ramp storage capacity:
\begin{equation}
\hat{v}_{z,t}=\frac{\bar{v}_{z,t}}{v_{\mathrm{ff}}},
\qquad
\hat{o}_{z,t}=\frac{o_{z,t}}{o_{\max}},
\qquad
\hat{l}_{q,t}=\frac{l_q(t)}{l_{q,\max}}.
\end{equation}
The larger weights assigned to merge speed and occupancy emphasize conditions at the bottleneck, while the queue penalty discourages excessive ramp accumulation.

A spillback-related penalty activates when the ramp queue exceeds a critical threshold $l_{q,\text{crit}}=0.9l_{q,\text{max}}$. The penalty increases linearly above this threshold:
\begin{equation}
C_{\text{spillback}} =
\begin{cases}
\frac{l_q(t) - l_{q, \text{crit}}}{l_{q, \text{max}} - l_{q, \text{crit}}} & \text{if } l_q(t) > l_{q, \text{crit}} \\
0 & \text{otherwise}
\end{cases}
\end{equation}
The spillback-related term is a soft penalty: its high weight discourages queues approaching ramp capacity but does not guarantee avoidance of spillback. The same reward formulation and weights are used across the DRL state-representation variants. The weights are summarized in Table~\ref{tab:hyperparams}.

\section{EXPERIMENTAL SETUP}
\label{sec:exp}

All experiments are conducted in SUMO \cite{Lopez2018} on a simulated 1.2\,km freeway bottleneck with a single-lane on-ramp, shown in Fig.~\ref{fig:network_schematic}. The network captures the interaction between mainline congestion and ramp queue accumulation.

SUMO network specifications, neural network architecture, and training hyperparameters are summarized in Tables~\ref{tab:network_specs},~\ref{tab:network_arch} and~\ref{tab:hyperparams}, respectively.

The implementation builds on the open-source framework proposed in \cite{DucrocqFarhi2023}, which provides a stable integration between SUMO and DRL algorithms for traffic control. We adapt this framework to the freeway ramp metering context by introducing the hybrid observation representation, the Two-Stream neural network, and a revised state-action interface. Interaction between the agent and the simulator is handled through the TraCI library.

\begin{table}[htbp]
    \caption{Action Mapping for Green-Phase Durations}
    \label{tab:action_map}
    \begin{center}
    \small
    \begin{tabular}{l cccccccc}
    \toprule
    \textbf{Action index $a_t$} & 0 & 1 & 2 & 3 & 4 & 5 & 6 & 7 \\
    \midrule
    \textbf{Green time $T_g$ (s)} & 5 & 10 & 15 & 20 & 25 & 30 & 35 & 40 \\
    \bottomrule
    \end{tabular}
    \end{center}
\end{table}

Training uses a stochastic curriculum to improve robustness. Each training episode simulates one hour of traffic. Mainline and on-ramp demands are sampled from discrete distributions over 4000--6500\,veh/h and 1400--2000\,veh/h, respectively, with greater probability assigned to congested conditions. At episode start, CV penetration is sampled as $p_{cv}\sim U(0.01,0.99)$.

All DRL architectures are trained using the same ordered sequence of episode-level scenario seeds and the same training budget. These seeds control the sampled demand and penetration conditions, providing matched scenario exposure across architectures. Each architecture undergoes one continuous training run containing multiple episodes. Variation in episode seeds therefore provides diverse training conditions but does not constitute independent repetition of the complete learning process.

The hybrid policy is trained under randomly varying CV penetration and evaluated under two observation conditions: partial CV visibility and full CV visibility ($p_{cv}=1$). Both conditions use the same trained network parameters, without retraining. The full-visibility condition serves as a reference for assessing the effect of incomplete microscopic observation. The proposed agent is compared to classical and learning baselines. Classical baselines include no-control (no ramp metering) and ALINEA \cite{papageorgiou1991}. Learning ablations include a DRL agent that uses macroscopic inputs only and a DRL agent with lane-level macroscopic inputs but without the microscopic grid.

Performance is measured over 50 one-hour evaluation scenarios. All controllers use the same seeded scenario stream to enable episode-by-episode comparisons. The full-CV-visibility condition uses the same trained hybrid policy on the corresponding traffic scenarios with $p_{cv}=1$. Reported means and standard deviations describe variation across evaluation scenarios for the trained policies, rather than variability across independent training runs. The reported metrics are total travel time, total time loss, total waiting time, spillback duration, and CO$_2$ emissions.
\section{RESULTS AND DISCUSSION}
\label{sec:res}

Table~\ref{tab:performance_summary} summarizes performance across the 50 matched evaluation scenarios.

\begin{table*}[t] \caption{Aggregated Performance Metrics Across Control Strategies (Mean $\pm$ Std. Dev.). Bold and underlined values indicate the lowest and second-lowest means, respectively, across all listed controllers.} \label{tab:performance_summary} \begin{center} \small \resizebox{\textwidth}{!}{ \begin{tabular}{lccccc} \toprule \textbf{Strategy} & \textbf{Travel Time (s)} & \textbf{Time Loss (s)} & \textbf{Wait Time (s)} & \textbf{Spillback (s)} & \textbf{CO$_2$ ($10^9$ mg)} \\ \midrule No Control & $577,190 \pm 72,494$ & $338,777 \pm 69,578$ & $28,712 \pm 6,539$ & $\mathbf{1.48 \pm 4.04}$ & $2.11 \pm 0.15$ \\ ALINEA & $421,540 \pm 41,909$ & $170,540 \pm 34,937$ & $19,711 \pm 5,514$ & $455.7 \pm 206.4$ & $1.87 \pm 0.12$ \\ \midrule DQN Macro (No Lane) & $382,884 \pm 35,719$ & $130,092 \pm 26,793$ & $15,399 \pm 6,771$ & $94.40 \pm 105.6$ & $1.79 \pm 0.12$ \\ DQN Macro + Lane & $376,819 \pm 31,966$ & $124,718 \pm 24,142$ & $13,437 \pm 4,737$ & $44.00 \pm 64.7$ & \underline{$1.78 \pm 0.10$} \\ \textbf{DQN Hybrid (Partial)} & \underline{$373,633 \pm 29,646$} & \underline{$120,256 \pm 21,292$} & \underline{$12,404 \pm 4,593$} & $68.80 \pm 95.6$ & \underline{$1.78 \pm 0.10$} \\ \textit{DQN Hybrid (Full)} & $\mathbf{367,825 \pm 28,766}$ & $\mathbf{112,530 \pm 19,319}$ & $\mathbf{12,345 \pm 4,522}$ & \underline{$37.60 \pm 53.2$} & $\mathbf{1.77 \pm 0.10}$ \\ \bottomrule \end{tabular} } \end{center} \end{table*}

\subsection{Overall Control Performance}

The reported mean travel time decreases as the observation representation becomes richer. The hybrid controller under partial CV visibility achieves $373{,}633$\,s, compared with $421{,}540$\,s for ALINEA, corresponding to an $11.4\%$ reduction. Relative to the lane-level macroscopic controller, the reduction from $376{,}819$\,s to $373{,}633$\,s is approximately $0.85\%$. This incremental improvement is smaller than the improvement relative to ALINEA. The ordering of the reported means alone does not establish the statistical significance of differences between the DRL representations.

Under full CV visibility, the same trained hybrid policy achieves a reported total travel time of $367{,}825$\,s and time loss of $112{,}530$\,s. The partial-visibility travel time is approximately $1.6\%$ higher than this reference. This comparison measures the performance difference associated with observation availability for the fixed learned policy, rather than a difference between separately trained controllers.

The partial-visibility hybrid controller records $68.8$\,s of mean spillback duration, compared with $455.7$\,s for ALINEA, a reduction of approximately $84.9\%$. However, the lane-level macroscopic controller records a lower mean spillback duration of $44.0$\,s. The hybrid controller therefore does not dominate this baseline on every metric.

No control produces the lowest ramp spillback duration but the highest reported travel time. This illustrates the ramp-metering trade-off: restricting ramp inflow can improve mainline conditions while increasing ramp queue accumulation. The lower travel time and higher spillback duration of the partial-visibility hybrid controller relative to the lane-level baseline are consistent with this trade-off.

\begin{figure}[t]
    \centering
    \includegraphics[width=\linewidth]{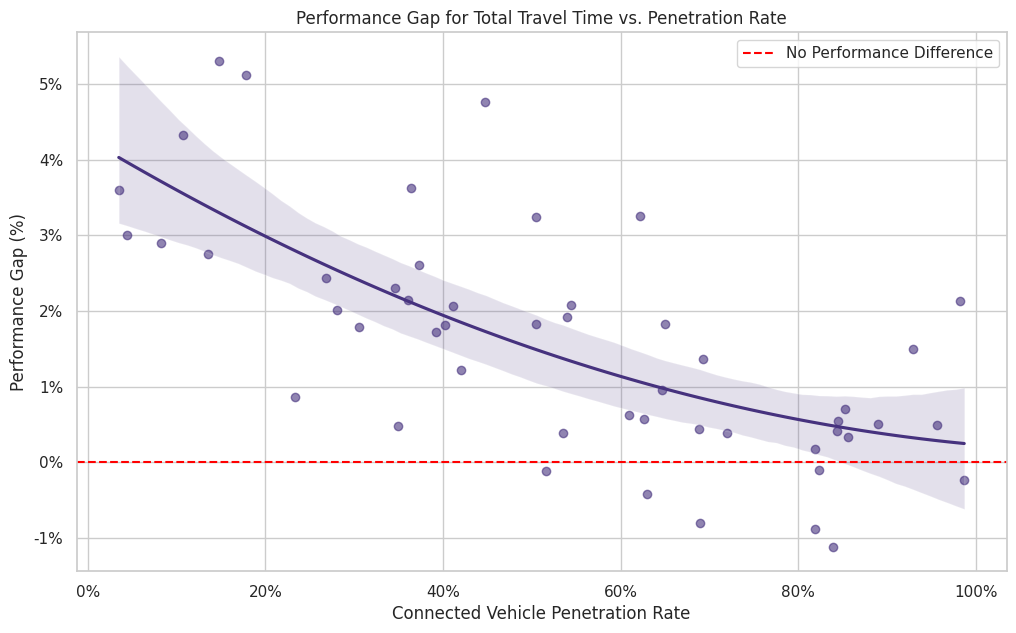}
    \caption{Relative travel-time gap between partial and full CV visibility as a function of CV penetration for the same trained hybrid policy. The horizontal dashed line indicates equal performance.}
    \label{fig:penetration_sensitivity}
\end{figure}

\subsection{Environmental Performance and Variability}

Mean CO$_2$ emissions are $1.78\times10^9\,\mathrm{mg}$ under partial hybrid visibility and $1.77\times10^9\,\mathrm{mg}$ under full hybrid visibility. At the reported precision, the partial-visibility hybrid controller and the lane-level macroscopic controller have the same mean emissions. The reported standard deviations describe scenario-to-scenario variability and should not be interpreted as evidence of closed-loop stability.

\subsection{Effect of CV Penetration}

Fig.~\ref{fig:penetration_sensitivity} examines how the travel-time gap relative to full CV visibility varies with penetration for the same trained hybrid policy. The fitted trend suggests that the gap decreases as penetration increases, with smaller additional improvements at higher penetration rates. Around $60\%$ penetration, the fitted gap is approximately $1\%$ in the evaluated scenarios.

This relationship is descriptive rather than a universal penetration threshold. Traffic conditions also vary across the plotted scenarios, so the fitted curve should not be interpreted as a controlled comparison at each fixed penetration level. Negative gaps in individual scenarios indicate that full CV visibility does not guarantee better performance for every realization of the fixed learned policy.

\begin{figure}[t]
    \centering
    \includegraphics[width=\linewidth]{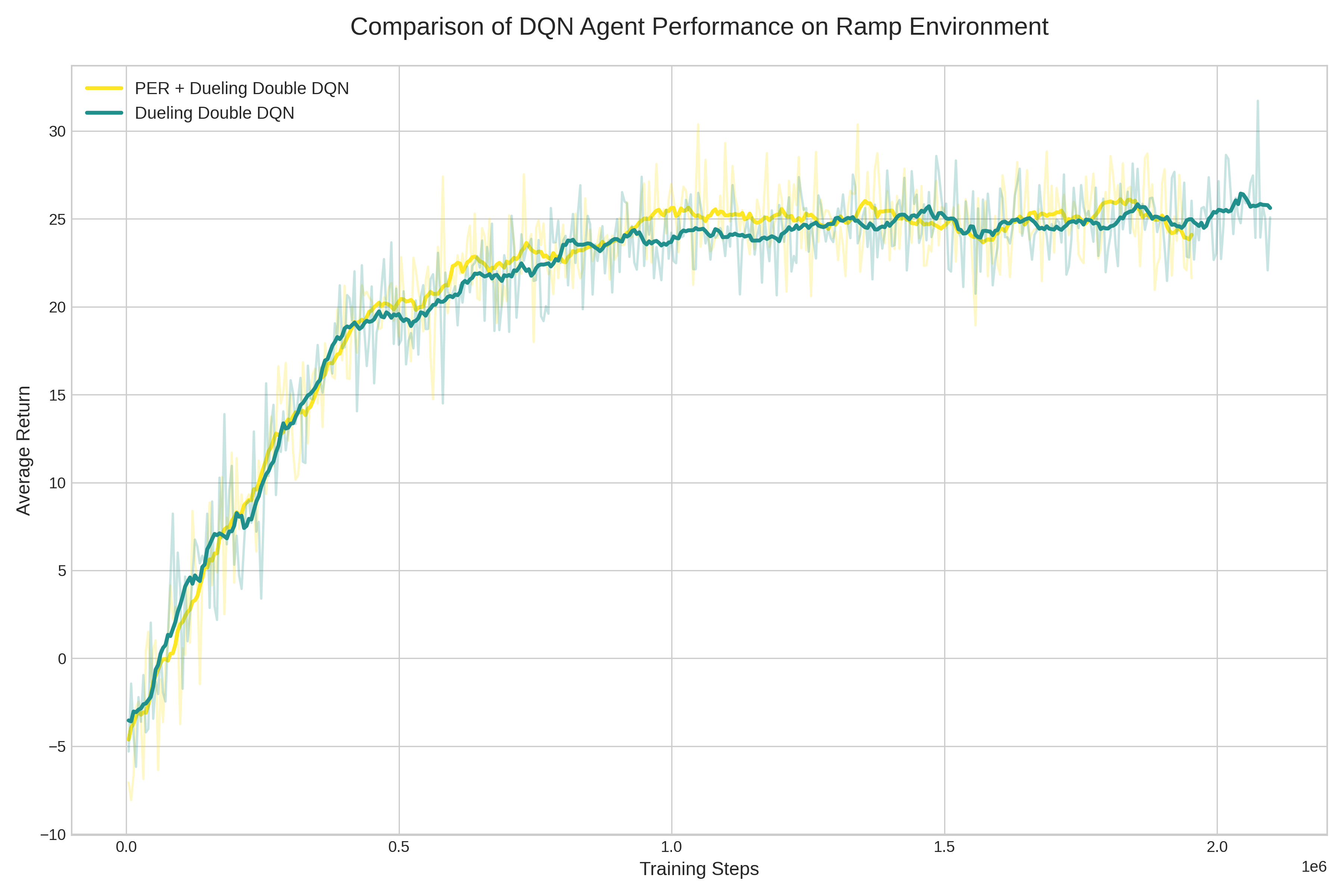}
    \caption{Training-return trajectory of the hybrid Dueling Double DQN controller.}
    \label{fig:training_curve}
\end{figure}

\subsection{Learning Behavior}

Fig.~\ref{fig:training_curve} shows the training-return trajectory of the hybrid controller. Returns increase during the initial learning phase and subsequently approach a plateau. This provides evidence of improved training performance in the observed run but does not establish convergence to an optimal policy or consistency across independent training runs.

\subsection{Limitations}

The evaluation concerns a single simulated freeway bottleneck and the specified traffic-demand distribution. The results do not establish transferability to other network geometries, driving behaviors, or sensing conditions. Communication delays and measurement errors are not evaluated. Furthermore, the reported variability is across traffic scenarios for the trained policies, rather than across independent training runs. Matching scenario seeds controls experimental conditions across controllers but does not establish reproducibility under different initializations and learning randomness.


\section{CONCLUSIONS}
\label{sec:conc}

This paper investigated learning-based ramp metering using complementary macroscopic measurements and sparse connected-vehicle observations. The proposed hybrid representation combines aggregate traffic information with a spatial grid of observed vehicle presence and speed, processed by a Dueling Double DQN controller.
Across 50 matched SUMO evaluation scenarios, the hybrid controller under partial CV visibility reduced the reported mean total travel time by $11.4\%$ and mean spillback duration by $84.9\%$ relative to ALINEA. The travel-time improvement relative to the lane-level macroscopic DRL baseline was smaller, approximately $0.85\%$, and was accompanied by higher mean spillback duration.
Evaluating the same trained hybrid policy with full CV visibility reduced reported travel time by a further $1.6\%$. The penetration analysis suggests a decreasing performance gap as microscopic observations become more complete. These findings support the usefulness of combining aggregate measurements with incomplete vehicle-level observations within the evaluated setting.
Future work will assess variability across independent training runs, isolate penetration effects through matched evaluations at fixed visibility levels, and examine transfer to other networks and imperfect communication conditions.


\appendix

\section{Experimental specifications:}
SUMO network specifications, neural network architecture, and hyperparameters are reported in Tables~\ref{tab:network_specs}, \ref{tab:network_arch} and~\ref{tab:hyperparams} respectively.

 \begin{table}[h]
     \caption{Key Edge Specifications in the SUMO Network}
     \label{tab:network_specs}
     \centering
     \footnotesize
     \setlength{\tabcolsep}{3pt}
     \begin{tabular}{llccc}
     \toprule
     \textbf{Edge ID} & \textbf{Description} & \textbf{Length (m)} & \textbf{Lanes} & \textbf{Speed (m/s)} \\
     \midrule
     \texttt{main\_road}    & Upstream mainline    & 488.3 & 3 & 27.78 \\
     \texttt{on\_ramp}      & Ramp to signal       & 204.4 & 1 & 13.89 \\
     \texttt{passage\_area} & Signal to merge      & 42.5  & 1 & 13.89 \\
     \texttt{accel\_area}   & Merge / acc. lane    & 193.8 & 4 & 22.22 \\
     \texttt{end\_main}     & Downstream section   & 193.1 & 3 & 27.78 \\
     \texttt{off\_ramp}     & Diverging off-ramp   & 161.3 & 2 & 13.89 \\
     \bottomrule
     \end{tabular}%
 \end{table}

\begin{table}[h]
    \caption{Two-Stream Neural Network Architecture}
    \label{tab:network_arch}
    \centering
    \footnotesize
    \setlength{\tabcolsep}{4pt}
    \begin{tabular}{llc}
    \toprule
    \textbf{Component} & \textbf{Specification} & \textbf{Activation} \\
    \midrule
    Microscopic (CNN)  & Conv2D (32, 64, 64 filters)       & ELU \\
    Macroscopic (FC)   & Input: 14 features $\rightarrow$ FC(128) & ELU \\
    Shared Dense       & FC(512) $\rightarrow$ FC(256)      & ELU \\
    Value Head         & FC(256) $\rightarrow$ FC(1)        & Linear \\
    Advantage Head     & FC(256) $\rightarrow$ FC(8)        & Linear \\
    \bottomrule
    \end{tabular}%
\end{table}

\begin{table}[h]
    \caption{DRL Hyperparameters and Reward Weights}
    \label{tab:hyperparams}
    \centering
    \footnotesize
    \setlength{\tabcolsep}{3pt}
    \begin{tabular}{lc @{\hspace{8pt}} lc}
    \toprule
    \textbf{Hyperparameter} & \textbf{Value} & \textbf{Reward Weight} & \textbf{Value} \\
    \midrule
    Total training steps    & $2.1 \times 10^6$ & $w_{\text{speed\_merge}}$ (merge speed) & 1.5 \\
    Learning rate ($\alpha$)& $1 \times 10^{-4}$ & $w_{\text{speed\_up}}$ (up speed)        & 1.0 \\
    Discount factor ($\gamma$) & $0.99$         & $w_{\text{speed\_down}}$ (down speed)    & 0.5 \\
    Mini-batch size         & 32                & $w_{\text{occ\_bottle}}$ (merge occ.)   & 2.0 \\
    Replay buffer size      & $1 \times 10^6$   & $w_{\text{occ\_up}}$ (up occ.)          & 1.0 \\
    Target update ($\tau$)  & $10^{-3}$         & $w_{\text{queue}}$ (queue penalty)      & 1.0 \\
    Exploration ($\epsilon$)& $1.0 \rightarrow 0.01$ & $w_{\text{spillback}}$ (spillback) & 20.0 \\
    \bottomrule
    \end{tabular}
\end{table}

\end{document}